\documentclass[sigconf]{acmart}
\renewcommand\footnotetextcopyrightpermission[1]{} 
\usepackage{booktabs} 

\setcopyright{rightsretained}
\usepackage{mathtools}
\usepackage{graphicx}
\usepackage{amsmath,amssymb} 
\usepackage{color}
\usepackage{subcaption}
\graphicspath{{data/}}

\acmDOI{10.475/123_4}

\acmISBN{123-4567-24-567/08/06}

\acmArticle{4}
\acmPrice{15.00}

\editor{Jennifer B. Sartor}
\editor{Theo D'Hondt}
\editor{Wolfgang De Meuter}

\begin{document}
\title{Video Captioning with Boundary-aware Hierarchical Language Decoding and Joint Video Prediction} 
\author{Xiangxi Shi}
\affiliation{%
	\institution{Nanyang Technological University}
	\city{Singapore}
	\state{Singapore}
}
\email{xxshi@ntu.edu.sg}
\author{Jianfei Cai}
\affiliation{%
	\institution{Nanyang Technological University}
	\city{Singapore}
	\state{Singapore}
}
\email{asjfcai@ntu.edu.sg}
\author{Jiuxiang Gu}
\affiliation{%
	\institution{Nanyang Technological University}
	\city{Singapore}
	\state{Singapore}
}
\email{JGU004@e.ntu.edu.sg}
\author{Shafiq Joty}
\affiliation{%
	\institution{Nanyang Technological University}
	\city{Singapore}
	\state{Singapore}
}
\email{srjoty@ntu.edu.sg}


\begin{abstract}
The explosion of video data on the internet requires effective and efficient technology to generate captions automatically for people who are not able to watch the videos. Despite the great progress of video captioning research, particularly on video feature encoding, the language decoder is still largely based on the prevailing RNN decoder such as LSTM, which tends to prefer the frequent word that aligns with the video. In this paper, we propose a boundary-aware hierarchical language decoder for video captioning, which consists of a high-level GRU based language decoder, working as a global (caption-level) language model, and a low-level GRU based language decoder, working as a local (phrase-level) language model. Most importantly, we introduce a binary gate into the low-level GRU language decoder to detect the language boundaries. Together with other advanced components including joint video prediction, shared soft attention, and boundary-aware video encoding, our integrated video captioning framework can discover hierarchical language information and distinguish the subject and the object in a sentence, which are usually confusing during the language generation. Extensive experiments on two widely-used video captioning datasets, MSR-Video-to-Text (MSR-VTT) \cite{xu2016msr} and YouTube-to-Text (MSVD) \cite{chen2011collecting} show that our method is highly competitive, compared with the state-of-the-art methods.
\end{abstract}

%
%


\maketitle

\section{Introduction}
With the explosion of mass video data such as surveillance videos and personal video data, it is highly desired to develop automatical video understanding technology. Early video understanding tasks such as scene understanding and action recognition focus on generating a few labels for videos. To further comprehend video content, video captioning becomes a popular topic in computer vision, which aims to generate natural descriptions from videos. A similar task is image captioning, which tries to generate natural descriptions from images. Both image captioning and video captioning are challenging tasks since it requires the knowledge of both language processing and visual content processing. On one hand, image elements such as scenery and objects in visual content needs to be captured and represented. On the other hand, language rules such as text context should be applied to generate meaningful sentences.

Early video captioning works were typically developed as an extension of image captioning \cite{vinyals2015show,karpathy2015deep}, aiming to linking video elements to words. Compared with an image, a video contains much more information including not only object and scene but also action and interactions. To better encode video information, many methods have been proposed. In particular, Resnet \cite{he2016deep} and 3D ConvNets \cite{tran2015learning} have been introduced into video captioning to summarize motion information in short videos. The Long Short-Term Memory (LSTM) \cite{hochreiter1997long}, which has been proved successful in natural language processing (NLP), has also been applied to capture the temporal information in video in addition to the sentence generation. The recent work \cite{chung2014empirical} proved that some variant of Recurrent Neural Network (RNN) like Gated Recurrent Unit (GRU) \cite{cho2014learning,kang2016gated} are more suitable for short-sequence data because of its simple structure with fewer parameters, which is easier to be optimized and relieved from overfitting.

Despite the great progress of video captioning, some problems still remain unsolved. In particular, most of the recent works focus on improving video encoding, while their language decoders are still based on the prevailing RNN decoder such as LSTM to learn the language dependency. Although a language dependency can help generate a more reasonable description, a too strong language dependency will lead to mistakes by linking two words that do not appear together in the video but appear together frequently in the dataset. For example, the word ``sheep" often appears with ``meadow", but a video could have meadow without sheep. In addition, when a word (e.g woman) appears in the dataset more frequently than another word (e.g. motorcycle), the language generator tends to predict the former than the latter when both occur in a video, results in a caption like ``a woman is riding a woman" instead of ``a woman is riding a motorcycle".

Therefore, in this paper we propose a boundary-aware hierarchical language decoder for video captioning. It consists of a high-level GRU based language decoder, working as a global (caption-level) language model, and a low-level GRU based language decoder, working as a local (phrase-level) language model. The key novelty lies in that we introduce a binary gate into the low-level GRU language decoder, named Binary Gated Recurrent Unit (B-GRU), to detect the language boundaries according to the language information and feed them back to the high-level language decoder to generate a global understanding of the currently generated sentence segments. Note that boundary detection has been introduced in \cite{baraldi2016hierarchical} but only for video encoding. In contrast, our method applies boundary detection concepts for not only video encoding but also language decoding. Moreover, to further improve the performance, we also incorporate another task of video prediction for multi-task learning as well as integrating shared attention model for both video captioning and video prediction tasks. Figure~\ref{fig:top} gives an overview of our video captioning framework.


In summary, our main contributions are threefold:
\begin{itemize} 
	\item We propose a novel boundary-aware hierarchical language decoder for video captioning.
	\item We develop a video captioning framework that integrates the proposed boundary-aware hierarchical language decoder with several advanced components including joint video prediction, shared soft attention, and boundary detection for video encoding. Such an integration is non-trivial. 
	\item We conduct extensive experiments on two widely-used video captioning datasets, MSR-Video-to-Text (MSR-VTT) \cite{xu2016msr} and YouTube-to-Text (MSVD) \cite{chen2011collecting}. Experimental results show that our method can achieve very competitive performance and produce better captions.
\end{itemize} 
\section{Related Work}
Video captioning is a hot topic in computer vision and natural language processing communities and many methods have been proposed.

\textbf{Template-based approaches.} Early works in video captioning~\cite{guadarrama2013youtube2text,krishnamoorthy2013generating,thomason2014integrating} treat the problem as a template-matching problem, focusing on generating video descriptions based on the identification of (subject, verb, object) triplets with visual classifiers.
However, such the template-based approaches have limited ability to generalize to unseen data, and the generated sentences cannot satisfy the richness of natural language.

\textbf{Encoder-decoder based methods.} Inspired by the success of deep neural networks in neural machine translation (NMT) and image captioning~\cite{vinyals2015show,karpathy2015deep},
the recent video captioning works have moved to utilize RNNs, which, given a vectored description of a visual content, can naturally deal with sequences of words~\cite{vinyals2015show,karpathy2015deep}.
The first work applying RNNs to video captioning is~\cite{venugopalan2014translating}, in which they proposed an encoder-decoder based framework for video caption generation.
They used Convolutional Neural Network (CNN) to extract features from each single frame, and took an average pooling over all the video frame features to get the entire video representation, which is then fed into an LSTM network for the sentence decoding.
However, their mean pooling operation ignores the sequential nature of videos.
After that, many works have tried to improve the video encoding process.
In \cite{donahue2015long}, they encoded the input video with another LSTM network and employed CRFs to generate a coherent video description.
Venugopalan et al.~\cite{venugopalan2015sequence} encoded a video with a stacked LSTM network, which was later improved by incorporating attention mechanisms~\cite{yao2015describing} in the sentence decoder, or combining with external knowledge~\cite{rohrbach2015long}.

\textbf{Other enhanced methods.} To generate more accurate captions, attribute detections are introduced into the model to detect semantic attributes of a frame. Each element in an attribute vector represents the possibility of a particular semantic attribute appearing in the frame. Pan et al. \cite{pan2017video} proposed a video captioning model combined with semantic attribute detection. They introduced a transfer unit to merge the temporal information with image attributes and video attributes. Later on, Gan et al. \cite{gan2017semantic} further improved the method by proposing an individual LSTM structure using the semantic information to guide the weight matrix in LSTM. With the new structure, the semantic information is involved in language decoding process rather than being merged with the hidden state to generate a fused vector. Although these methods perform well in video captioning, they cannot get good results by only training with a video captioning dataset, since the video captioning datasets usually do not contain enough data to train a good attribute detector. Thus, all these methods need to involve additional image datasets to train their image attribute detection network.

Recently, Pasunuru et al. \cite{pasunuru2017multi} proposed a video captioning framework which   requires multiple datasets for training. Particularly, they added a video prediction process and an entailment generation process into the video captioning task to learn better representations of both video and language. They involved the UCF-101 \cite{soomro2012ucf101} action videos dataset to train the video prediction model and the Stanford Natural Language Inference (SNLI) corpus to train the entitlement generation.

\textbf{Hierarchical structure based approaches.} Compared with the encoder-decoder structure, the hierarchical structure introduces multiple layers into the model to learn different information at different layers. In particular, Song et al. \cite{song2017hierarchical} proposed a hierarchical language decoder to generate a sentence considering both low-level visual information and high-level language context information. They introduced an adjusted component to merge the attended vector and the hidden state of the language decoder. In this way, the model involves more direct video information to generate the words whose attributes appear in the video. On the other hand, the video encoder can also be enhanced with a hierarchical structure. Pan et. al. \cite{pan2016hierarchical}
proposed a two-layer RNN video encoder to model the temporal information efficiently by controlling the length of the video information flow. Different from the stacked RNN in the sequence-to-sequence method~\cite{sutskever2014sequence}, they introduced a hierarchical video encoder with a variable encoding interval. The low-level LSTM filter encodes the video features in individual chunks, each of which contains the same amount of features. The high-level encoder encodes the final outputs of the LSTM filter to generate an encoded vector representing the global understanding of the video. With the LSTM filters, the model can fully use all the video frames instead of the conventional way of selecting several keyframes. Baraldi et. al. \cite{baraldi2016hierarchical} proposed a boundary-aware method for video encoding model, which can be view as an extension of Pan's work \cite{pan2016hierarchical}. Specifically, they introduced a boundary detection into LSTM to automatically decide how long a chunk should be.

In contrast, our approach is a non-trivial integration of many advanced components including hierarchical language decoder, joint video prediction, attention mechanism, and boundary detection. Different from the approach in \cite{pasunuru2017multi}, which also incorporates joint video prediction, our method does not need multiple datasets and share the soft attention model between the video captioning process and the video prediction process. Different from the approach in \cite{baraldi2016hierarchical}, which uses hierarchical language decoder and boundary detection for the video encoder, our method incorporates boundary detection for both the video encoder and the hierarchical language decoder.

\section{Method}

Let $\mathbf{f}=(f_0,\cdots, f_{N-1})$ be a sequence of video frames, where $f_i$ is the image at time step $i$, and $N$ is  the length of the sequence. The video captioning task is to generate a text $\hat{\mathbf{Y}}=\{\hat{Y}_0, \cdots, \hat{Y}_{T-1}\}$ that describes the video, where $\hat{Y}_t\in \mathcal{D}$ is a word predicted from a vocabulary $\mathcal{D}$, and $T$ denotes the number of words in the description. 

\begin{figure*}[ht]
	\centering
	\vspace{2mm}
	\includegraphics[width=0.85\linewidth]{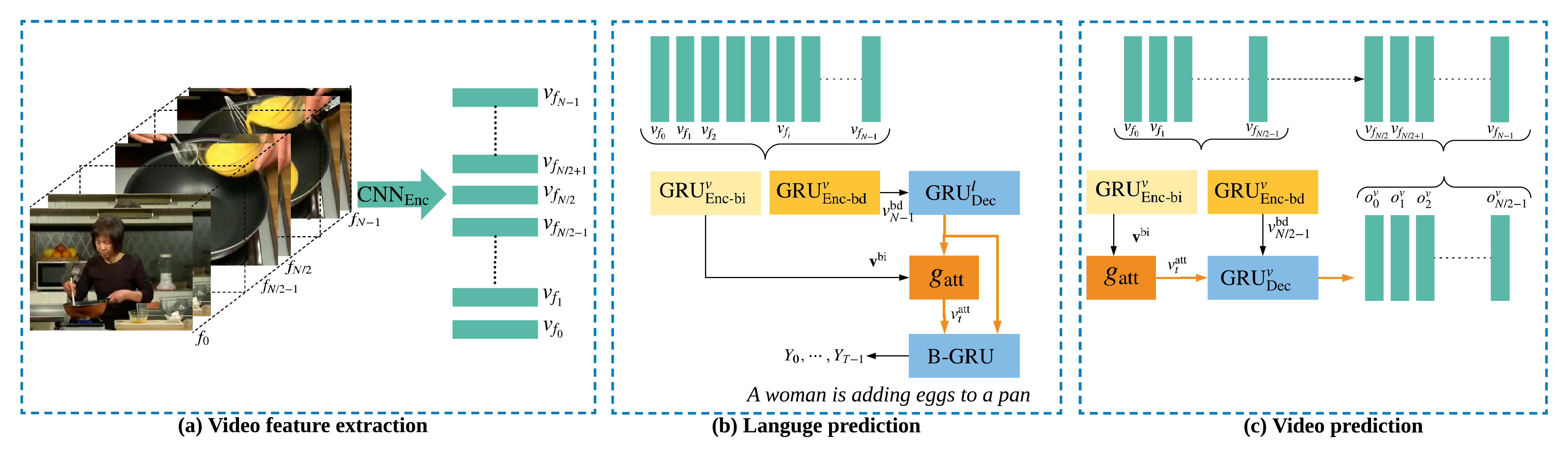} 
	\caption{Illustration of our proposed framework. Our model contains two training paths: one is for caption generation, and the other is for video prediction. Note that the parameter of two video encoder as well as the attention model are share across two training paths.}
	\label{fig:top}
\end{figure*}
Figure~\ref{fig:top} shows our proposed framework for video caption generation. It consists of three parts: (i) video feature extraction, (ii) language prediction, and (iii) video prediction. As shown in Figure~\ref{fig:top}(a), a pre-trained CNN is used to encode each  video frame $f_i \in \mathbf{f}$ into a feature vector $v_{f_i}$ yielding video features $\mathbf{v}_f=(v_{f_0},\cdots, v_{f_{N-1}})$. Specifically, we obtain the image features $v_{f_i}$ by performing a mean-pooling over the spatial image features in the final convolutional layer of a CNN that is pre-trained on a large image recognition dataset like ImageNet \cite{deng2009imagenet}. The extracted image features are then fed into an RNN encoder-decoder framework that generates a caption by encoding the image features sequentially. In the following, we first describe a simple encoder-decoder framework as a baseline reference for video captioning. Then we present our proposed multi-task learning approach with improved attention mechanism.





\subsection{Baseline Video Caption Generation Model}\label{sec:baseline}

\begin{figure}[htbp]
	\centering
	\includegraphics[width=\linewidth]{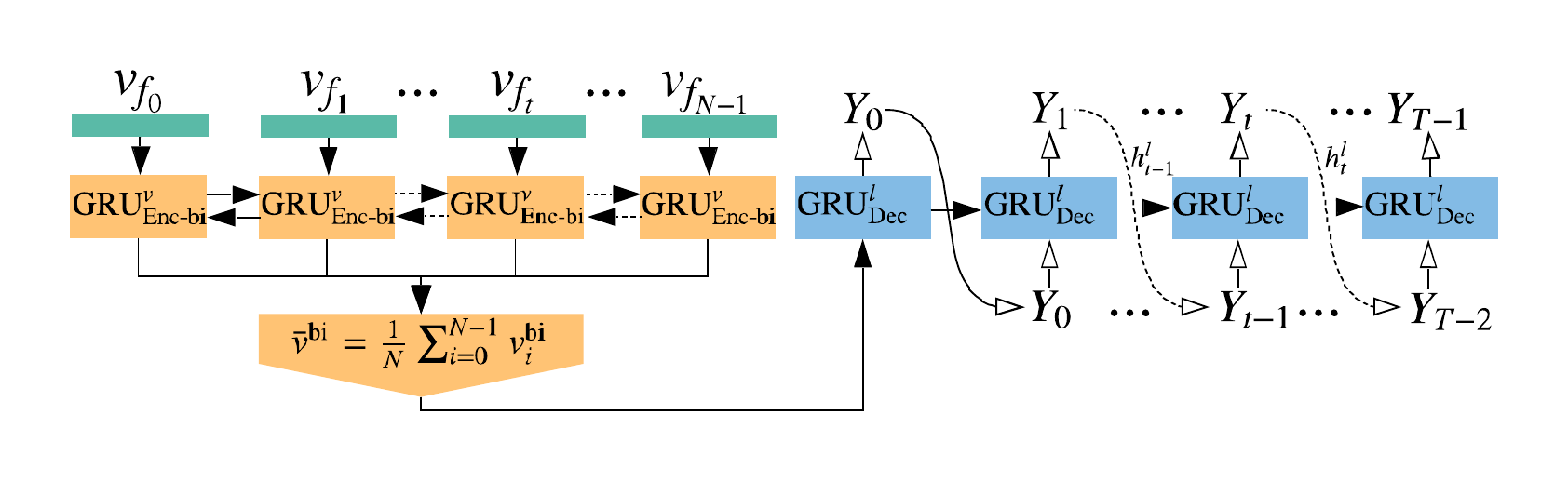} 
	\caption{Our encoder-decoder baseline with a bi-directional GRU encoder.}
	\label{fig:baseline}
\end{figure}

Our baseline video captioning model (Figure \ref{fig:baseline}) is similar to the standard encoder-decoder RNN model used in machine translation \cite{sutskever2014sequence}. To model the temporal information in
a video, we employ a bi-directional RNN with GRU recurrent cells~\cite{cho2014learning} as the encoder. The operations of the encoder ($\text{GRU}_{\text{Enc-bi}}^v$) can be expressed as:

\begin{equation}
(v_{0}^{\text{bi}},\cdots, v_{N-1}^{\text{bi}}) = \text{GRU}_{\text{Enc-bi}}^v ([v_{f_0},\cdots, v_{f_{N-1}}]) \label{eq:gru_base}\\
\end{equation}

\noindent where $v_i^{\text{bi}}$ is the concatenated output of the bidirectional GRU video encoder at time step $i$. Like LSTMs \cite{hochreiter1997long}, GRUs are good at memorizing long sequences, but computationally more efficient.

The decoder $\text{GRU}_{\text{Dec}}^l$ is a unidirectional RNN with GRU cells, which is fed with the \emph{average} of the encoded vectors  $\{v_{0}^{\text{bi}},\cdots, v_{N-1}^{\text{bi}}\}$  as the input at the first time step ($t=0$). The decoding step at time step $t$ can be expressed as: 

\begin{equation}
h_t^l, o_t^l = \text{GRU}_{\text{Dec}}^l(h_{t-1}^l, y_{t-1})\\
\end{equation}
\begin{equation}
Y_t\sim \text{argmax}(\mathbf{W}_\text{o}^\text{l} o_t^{\text{l}}+\mathbf{b}_\text{o}^\text{l})
\end{equation}
where $h_t^l$ is the hidden state of $\text{GRU}_{\text{Dec}}^l$, $o_t^l$ is the output of $\text{GRU}_{\text{Dec}}^l$. $Y_t$ is the index of rhe predicted word which has the highest probability in the Softmax process, $y_{t-1}=\mathbf{W}_eY_{t-1}$ is the embedding of previous word $Y_{t-1}$, and $Y_t$ is a word index drawn from the dictionary according to the output probability of Softmax.

The baseline model has two major problems. First, the language decoder generates words based on a single global video feature vector (i.e., $\bar{v}^{\text{bi}} = avg(v_{0}^{\text{bi}},\cdots, v_{N-1}^{\text{bi}})$). However, since a caption describes the whole video, some words in the caption only relate to a specific portion of the video. For example, consider the sequence of actions in Figure~\ref{fig:top}. The textual entity \emph{woman} is present (as an image object) only in the first few image frames, and the entity \emph{egg} comes only in the later part of the video. Decoding words based on a single global video representation could lead to sub-optimal results due to irrelevant information in the encoded feature vector, especially when the video sequence is long.


The second problem is that the simple GRU-based video encoder lacks sufficient temporal supervision to be able to correctly capture the event sequence in a video. Ideally, we would want the video encoder to learn representations that are useful for not only caption generation, but also to maintain the temporal information in videos \cite{pasunuru2017multi}. In the following, we present our approach to address these limitations of the baseline model.

\subsection{Our Approach}

To focus on the relevant frames of a video while decoding caption words, we propose a novel hierarchical language decoder with global visual attention. Additionally, we propose to use a boundary-aware video encoder \cite{baraldi2016hierarchical} to separate out the actions (e.g., `woman handling a pan' and `adding eggs to the pan' in Figure~\ref{fig:top}) in a video. 
To better capture temporal information in the encoded video features, we include an additional training path where we learn representations by generating a portion of the video sequence. In the following, we describe our proposed methods in detail.


\subsubsection{Hierarchical Language Decoding with Visual Attention}

Figure~\ref{fig:top}(b) illustrates our caption generation model for a given sequence of image features $\mathbf{v}_f=(v_{f_0},\cdots, v_{f_{N-1}})$. The spatial image features are first encoded into a temporal sequence of feature representations $(v_{0}^{\text{bi}},\cdots, v_{N-1}^{\text{bi}})$ with a GRU-based RNN  following Eq.~\eqref{eq:gru_base}. The model then uses a hierarchical language decoder with a visual attention layer to selectively attend to encoded video features $\mathbf{v}^{\text{bi}} =\{v_{0}^{\text{bi}},\cdots, v_{N-1}^{\text{bi}}\}$ while generating the caption words. 

\begin{figure}[t]
	\centering
	\includegraphics[width=0.85\linewidth]{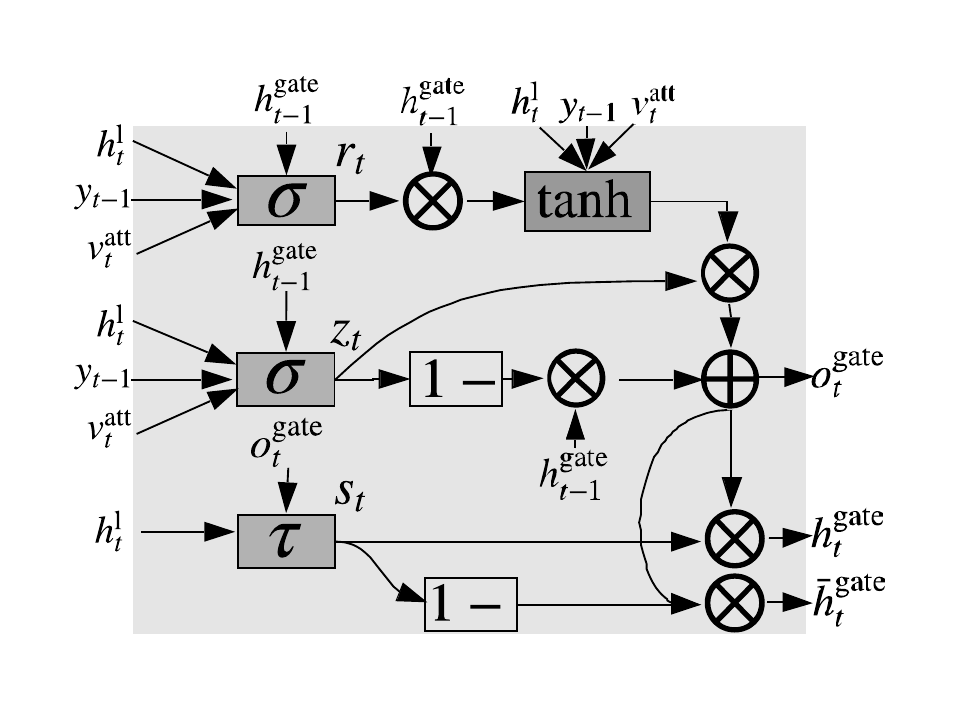}
	\caption{The structure of our proposed Binary Gated Recurrent Unit.}\label{fig:bgru}
	\vspace{-2mm}
\end{figure}
Figure~\ref{fig:top}(b) shows our proposed language decoder, specifically designed for video captioning. The decoder comprises two intertwined GRU layers. {The GRU at the top ($\text{GRU}_{\text{Dec}}^l$) works as a global (caption-level) language model, while the GRU at the bottom ($\text{B-GRU}$) works a local (phrase-level) language model. The B-GRU generates words to form consistent phrases (e.g., `A woman', `adding eggs in a pan') by attending on the relevant portion of the video, while the global GRU guides the B-GRU to generate globally consistent sentences.}  


The inputs to $\text{GRU}_{\text{Dec}}^l$ consist of its previous hidden state $h_{t-1}^l$ and $\bar{h}_{t-1}^{\text{gate}}$ of the $\text{B-GRU}$: 
\begin{align}
h_t^l&= \text{GRU}_{\text{Dec}}^l(\bar{h}^{\text{gate}}_{t-1},h_{t-1}^l)\label{eq:lang0}
\end{align}

We use the hidden state $h_{t}^l$ at each time step $t$ to generate an attention (or context) vector as follows: 

\begin{equation}
{v_t^{\text{att}} = g_{\text{att}}(\mathbf{v}^{\text{bi}},h_{t}^{l})    =        \sum_{n=0}^{N-1} \alpha^t_n \cdot v^{\text{bi}}_{n} }
\label{eq:att_0}
\end{equation}
where $\alpha^t_n$ is the attention weight computed as follows:

\begin{align}
A^t_{n} &= \text{tanh}(\mathbf{W}_h^e h_{t}^{l}  + \mathbf{W}^e_v v_{n}^{\text{bi}}) \\
\mathbf{\alpha^t}&= \text{softmax}(\mathbf{W}_{a} \mathbf{A}^{t}+\mathbf{b}_{a})
\end{align}
with $\mathbf{W}_h^e$, $\mathbf{W}^e_v$, and $\mathbf{W}_{a}$ being the weight matrices, and $\mathbf{b}_{a}$ being the bias vector.


The input to the B-GRU consists of its previous hidden state $h^{\text{gate}}_{t-1}$, the attention vector $v_t^{\text{att}}$, the updated hidden state $h^l_{t}$, and the embedding of the previous word $y_{t-1}$:

\begin{equation}
o^{\text{gate}}_t, h^{\text{gate}}_t, \bar{h}^{\text{gate}}_t = \text{B-GRU}\left(v_t^{\text{att}}, h^l_{t}, h^{\text{gate}}_{t-1}, y_{t-1}\right)\label{eq:lang1}
\end{equation}

Finally, the output layer of the decoder RNN predicts the words:
\begin{equation}
Y_t \sim \text{argmax}(\mathbf{W}_\text{o}^\text{l} o_t^{\text{gate}}+\mathbf{b}_\text{o}^\text{l})
\end{equation}
where $\mathbf{W}_\text{o}^\text{l}$ is the weight matrix and $\mathbf{b}_\text{o}^\text{l}$ is the bias vector at the output layer, and $Y_t$ is the index of the predicted word which has the highest probability. In the following, we describe our proposed binary GRU unit in detail. 


\paragraph{\textbf{Binary Gated Recurrent Unit (B-GRU)}}

 The B-GRU layer works as a local language model, and is responsible for generating words that constitute consistent phrases. Figure~\ref{fig:bgru} shows the structure of a B-GRU cell. The transformation functions can be described as:

\begin{align}
z_t &= \sigma(\mathbf{W}_z[h_t^l;y_{t-1};v_t^{att};h^{\text{gate}}_{t-1}]+\mathbf{b}_z)\label{eq:9}\\
r_t &= \sigma(\mathbf{W}_r[h_t^l;y_{t-1};v_t^{att};h^{\text{gate}}_{t-1}]+\mathbf{b}_r)\\
\tilde{h}_t&=\tanh(\mathbf{W}[r_t \cdot h^{\text{gate}}_{t-1}; h_t^l;y_{t-1};v_t^{att}]) \\
{o}^{\text{gate}}_t &=  z_t \cdot\tilde{h}_t + (1 - z_t) \cdot h^{\text{gate}}_{t-1}\\
s_t &=\tau \left( \sigma ( \mathbf{W}_s {o}^{\text{gate}}_{t} + \mathbf{W}_m h_t^l +\mathbf{b}_s) \right)\\
{h}^{\text{gate}}_t &=  s_t \cdot {o}^{\text{gate}}_t\\
\bar{h}^{\text{gate}}_t& = (1 - s_t) \cdot {o}^{\text{gate}}_t
\label{eq:16}
\end{align}
where $\sigma(\cdot)$ is the sigmoid function, and $h^{\text{gate}}_{t-1}$ is the previous hidden state of the B-GRU. Similar to the original $\text{GRU}$ \cite{cho2014learning}, the B-GRU cell has a reset gate $r_t$ and an update gate $z_t$. The reset gate controls the forget rate of previous state $h^{\text{gate}}_{t-1}$ when computing the candidate activation $\tilde{h}_t$ for the current state. The update gate controls the contribution of $h^{\text{gate}}_{t-1}$ and $\tilde{h}_t$ in the candidate output $o^{\text{gate}}_t$.

To design B-GRU as a local language model, in addition to the standard update and reset gates, we introduce a new binary gate $s_t$, which generates a binary signal (i.e., $s_t \in \{ 0, 1 \}$) by comparing the local language information (${o}^{\text{gate}}_{t}$) with the global one ($h_t^l$). The step function $\tau(.)$ is the defined as follows:
\begin{align}
\tau(\sigma(x)) = \begin{cases} 1, & \mbox{if } \sigma(x) > 0.5 \\ 0, & \mbox{otherwise} \end{cases}
\label{eq:deterministic_step}
\end{align}



When $s_t = 1$, it means that the newly predicted word should be in the same phrase with previous words, and it hidden state $h_t^{\text{gate}}$ gets updated with $o_t^{\text{gate}}$. When $s_t = 0$, it means that there is no strong relationship between the new predicted word and the previously predicted words. In this situation, $\bar{h}^{\text{gate}}_{t-1}$ is sent to $GRU_{Dec}^l$, and its hidden state is refreshed. {Since the B-GRU will refresh its hidden state when the binary gate detects the end of a phrase, an output of B-GRU $o^{\text{gate}}_{t-1}$ only contains the information from the end of the last phrase to the word the model predicts currently.



As we start decoding a sentence, the initial state of $\text{GRU}_{\text{Dec}}^l$ is initialized with the final state of the video encoder. The $\text{GRU}_{\text{Dec}}^l$ thus starts with the global understanding of the video. Since, at each decoding step the $\text{GRU}_{\text{Dec}}^l$ is encoded with the local language information ${o}^{gate}_{t-1}$ and its previous state (Eq. \ref{eq:lang0}), the output $h^l_t$ contains both the global video information and the local language model. We use $h^l_t$ to attend to the relevant encoded video features, and to generate an attention vector informed by both the vision and the language sides. The attention vector is then sent to the B-GRU layer as an input along with $h^l_t$, the previous state $h^{\text{gate}}_{t-1}$ and the current input embedding $y_{t-1}$ (Eq. \ref{eq:lang1}). The rational for feeding $h^l_t$ to B-GRU is that it gives necessary global language information when its memory is re-initialized using the binary gate $s_t$.

\subsubsection{Boundary-aware Video Encoder.}

To separate out the actions in a video, in addition to video encoding with attention, we also consider a boundary-aware video  encoder \cite{baraldi2016hierarchical}. We denote this boundary-aware encoder with $\text{GRU}_{\text{Enc-bd}}^v$. 


\begin{align}
\{v_{0}^{\text{bd}},\cdots, v_{N-1}^{\text{bd}}\}=&\text{GRU}_{\text{Enc-bd}}^v ([v_{f_0},\cdots, v_{f_{N-1}}])
\end{align}

\noindent The {transformation} functions for $\text{GRU}_{\text{Enc-bd}}^v$ are defined as follows:

\begin{align}
h_{t}^{\text{bd}_0} =& \text{GRU}_{\text{Enc-bd}}^{v0}(v_{f_t}, h_{t-1}^{\text{bd}_0})\\
\beta =& \tau \left( \sigma \left( f_{\text{bd}}  ({v_{f_t}},h_{t-1}^{\text{bd}_0}) \right) \right)  \\ 
h_t^{\text{bd}_0}=& h_t^{\text{bd}_0}\cdot \left(1-\beta \right)\\
h_{t}^{\text{bd}}, {v_t^{\text{bd}}} =& \text{GRU}_{\text{Enc-bd}}^{v1}(h_t^{\text{bd}_0} \cdot \beta, h_{t-1}^{\text{bd}})\label{eq:gru_bd_2}
\end{align}
\noindent where $t\in [0,N-1]$, $\tau(.)$ is an element-wise step function similarly defined as Eq. (\ref{eq:deterministic_step}), and  $f_{\text{bd}}(\cdot)$ is the boundary detection function, which maps the inputs into a one-dimensional vector by performing a linear transformation. 

The boundary detection mechanism ensures that the input data following a time boundary are not influenced by those seen before the boundary,  and generates a hierarchical representation of the video in which each chunk is composed of homogeneous frames. In our model, we take the final output as the initial input to both the decoders for video captioning and video prediction.

\subsubsection{Video Prediction}\label{sec:video_prediction}

To learn richer video representations that can  predict the temporal context and actions, we use an alternative (unsupervised) video prediction task 
as shown in Figure~\ref{fig:top}(c). The video encoder for this task shares the parameters with the video encoder ($\text{GRU}_{\text{Enc-bi}}^v$) for the caption generation task. 
For each training video, we divide the feature set into 2 subsets $ \{v_{f_0},\cdots, v_{f_{N/2-1}}\}$ and $ \{v_{f_{N/2}},\cdots, v_{f_{N-1}}\}$, which contain same number of contiguous frames.
The first subset is for video encoding, and the second subset is served as ground truth.

The prediction of each frame feature is defined as follows:
\begin{align}
h_{t}^v, o_t^{v} =\text{GRU}_{\text{Dec}}^v \left(g_{\text{att}}(\mathbf{v}^{\text{bi}}, h_{t-1}^{v} ),o_{t-1}^v, h_{t-1}^v \right) \label{eq:20}
\end{align}

\noindent where $h_{t-1}^{v}$ and $o_{t-1}^v$ are the hidden state and the predicted feature at the previous time step, 
 $g_{\text{att}}(\cdot)$ is the attention model which shares the same parameters with Eq.~(\ref{eq:att_0}). We take the final output of $\text{GRU}_{\text{Enc-bd}}^{v}$ as the initial step of the video decoder $\text{GRU}_{\text{Dec}}^v$. 

We train the two related tasks, video captioning and video prediction, jointly in a  multi-task learning framework \cite{luong2015multi}, and demonstrate that the captioning task benefits from incorporating complementary visual knowledge from the video prediction task. 



\subsection{Training}
Given the video features $\mathbf{v}_f=(v_{f_0},\cdots, v_{f_{N-1}})$ and the ground-truth caption $(Y_0, Y_1, ..., Y_{T-1})$, the language decoder is conditioned step by step on the first $t$ words of the caption (already generated) and on the corresponding video descriptors in $\mathbf{v}_f$, and is trained to produce the next caption word. We train the caption generation process by minimizing the cross entropy (XE) loss:

\begin{align}
\mathcal{L}_{\text{xe}}=-\sum_{t=0}^{T-1}\log p_{\theta_t}(Y_t|Y_{0:t-1},v_{f_{0:N-1}};\theta)
\label{eq:xe_loss}
\end{align}
where $p_{\theta_t}(Y_t|Y_{0:t-1},v_{f_{0:N-1}})$ is the output probability of the predicted word $Y_t$ with the model parameters $\theta$. We share the weights of the model across all time steps.

For the video prediction path, we optimize the model to minimize the mean squared errors (MSE):
\begin{align}
\mathcal{L}_{\text{MSE}}= \frac{2}{N} \sum_{n=0}^{N/2-1} \|o_n^{v} - v_{f_{N/2+n}}\|^2
\label{eq:mse_loss}
\end{align}
where $v_{f_{N/2+n}}$ serves as the ground truth frame features, and $o_n^{v}$ is the predicted frame features given by the video decoder. 
\section{Experiments}

\subsection{Datasets and Metrics}
\textbf{Microsoft Video Description Corpus (MSVD) \cite{chen2011collecting}.} MSVD contains 1,970 Youtube video clips, 85K English descriptions collected by Amazon Mechanical Turkers.
As done in previous works~\cite{guadarrama2013youtube2text,venugopalan2014translating}, we split the dataset into contiguous groups of videos by index number: 1,200 for training, 100 for validation and 670 for testing.

\textbf{MSR-Video to Text (MSR-VTT) \cite{xu2016msr}.} MSR-VTT is a large-scale video description dataset which contains 10K video clips collected from the Internet. These video clips last for 41.2 hours in total, covering the 20 representative categories and diverse visual content collected with 257 queries in the video engines. The dataset contains 200K clip-sentence pairs, and each clip is annotated with about 20 natural sentences. Compared to MSVD, MSR-VTT is more challenging, due to the large variety of videos.

\textbf{Evaluation Metrics}
We use three popular metrics for evaluation:
BLEU~\cite{papineni2002bleu}, METEOR~\cite{denkowski2014meteor}, CIDEr~\cite{vedantam2015cider}.
BLEU is a precision-based metric, calculated by matching the n-grams in candidate and reference captions.
METEOR evaluates the captions by matching n-grams in different forms such as surface, stem and paraphrase found in the generated caption and in references.
CIDEr measures consensus in video descriptions by performing a Term Frequency-Inverse Document Frequency weighting for each n-gram.

\subsection{Implementation Detail}
We sample 20 video frames at an equal interval for each video.
For each frame, we extract its frame feature with the Resnet101 network pre-trained on ImageNet~\cite{deng2009imagenet}.
We set the dimensionality of the frame features as 2048.

For the preprocessing of captions, we transform all letters in the captions to lowercase.
The corpus of the dataset contains many captions with variable lengths. The total time step needed for training depends on the longest sentence, and it affects the optimized process and the training time. For MSVD, we truncate all captions longer than 10 tokens.
For MSR-VTT, we set the maximum number of words to 15. At the end of a sentence, we add an end-of-sentence tag <EOS>. For a sentence shorter the maximum length, we fill <pad> signals in the elements with indexes larger than that of <EOS>.

During training, we use Adam \cite{kingma2014adam} to train our model. The learning rate is set to $1\times10^{-4}$.
We apply the greedy decoding in our training with a scheduled sampling ratio increasing from 0.0 to 0.1 by 0.025 at every 20 epochs.
The decay parameters are set as $\beta_1 = 0.9$ and $\beta_2 = 0.99$.
We incorporate a simple but efficient dropout layer following the RNN cells with a dropout rate of 0.5.
We add a beginning-of-sentence signal <BOS> as the initial word to start the language generation.
When a sentence reaches the end, <EOS> appears.

During testing, we also consider using beam search in our model. Beam search is the method that iteratively keeps the set of the $k$ best word sequences generated at time step $t$ as the candidates to generate next words. In our experiments, we find that when the beam size is six, our model gets the best performance in general. Particularly, we input <BOS> into the language decoder to start generating video descriptions. At each step, we choose the word with the maximum probability after softmax until we reach <EOS>.

\subsection{Baselines for Comparisons}
To get the insights, we consider the following variants of our method.
For a fair comparison, the output dimension of all hidden states are fixed to 512.

\textbf{BI:}
This is the simplest baseline, as described in Section~\ref{sec:baseline}, which is a conventional encoder-decoder video captioning model. Particularly, we use a bidirectional GRU as the video encoder and a one-layer GRU as the language decoder. This model is trained using Eq.~\eqref{eq:xe_loss}.

\textbf{BI + BD + SA:} This baseline use the same bidirectional GRU video encoder as BI, but with an additional boundary detection module (BD) and an attention module (SA).
For each time step, the sentence decoder $\text{GRU}_{\text{Dec}}^l$ predicts the next word based on the previous word and the attended frame features calculated by Eq.~\eqref{eq:att_0}.
Here, we initialize the hidden state of $\text{GRU}_{\text{Dec}}^l$ with the last output of $\text{GRU}_{\text{Enc-bd}}^v$.
We optimize this model using Eq.~\eqref{eq:xe_loss}.

\textbf{BI + BD + SA + VP:} This baseline introduces a video prediction (VP) process to BI+BD+SA.
The video encoder and the attention module of the video prediction path are shared with the video captioning process. We use the video decoder $\text{GRU}_{\text{Dec}}^v$ introduced in Section~\ref{sec:video_prediction}.
All hidden states are initialized with zeros, except the hidden state of $\text{GRU}_{\text{Dec}}^l$.
We optimize the two prediction processes alternatively according to the scheduled probability.
Particularly, in our experiments, we find that 0.5 is the best probability for the model performance because it can make the model learn the knowledge in the video side and the language side evenly.
During the video prediction training process, we freeze $\text{GRU}_{\text{Dec}}^l$, and optimize the network with Eq.~\eqref{eq:mse_loss}.
Similarly, we freeze the parameters of $\text{GRU}_{\text{Dec}}^v$ during the caption prediction process, and optimize the model with Eq.~\eqref{eq:xe_loss}.

\textbf{BI +BD + SA + VP + HLM:} This is the complete set of our method including all the components, where we improve the language decoder $\text{GRU}_{\text{Dec}}^l$ with a hierarchical language model (HLM). Here, the word decoder $\text{B-GRU}$ predicts the next word based on the previous word and the attended frame features as well as the hidden state of the global language model $\text{GRU}_{\text{Dec}}^l$. Different from BI+BD+SA+VP, the soft attention model here also takes in the output of $\text{GRU}_{\text{Dec}}^l$ instead of that of $\text{B-GRU}$. For this final model, we also consider using beam search during testing.

\begin{table}
	\begin{center}
		\small
		\caption{Comparisons of the video captioning results of different methods on the MSVD dataset.} 	\label{tab:msvd}
		\begin{tabular}{lp{1cm}p{1cm}p{1cm}} 
			\hline
			Model                                                                                 & BLEU 4     & METEOR         & CIDEr         \\
			\hline
			MM-VDN~\cite{xu2015multi}    & 34.0    & 29.0    & -            \\
			SA-GoogleNet+3D-CNN\cite{yao2015describing}    & 41.9    & 29.6    & -            \\
			S2VT\cite{venugopalan2015sequence}    & -        & 29.8    & -         \\
			LSTM-E\cite{pan2016jointly}    & 45.3    & 31.0    & -         \\
			Boundary-aware encoder\cite{baraldi2017hierarchical}    & 42.5    & 32.4     & 63.5         \\
			HRNE\cite{pan2016hierarchical}    & 44.8    & 33.1    & -     \\
			TDDF(VGG+C3D)\cite{zhang122017task}   &45.8   &\textbf{33.3}   &73.0\\
			\hline
			
			\text{our result}   & \textbf{50.3} & 32.9    &\textbf{74.3}         \\
			\hline
		\end{tabular}
	\end{center}
\end{table}

\begin{table}
	\begin{center}
		\small
		\caption{Comparisons of the video captioning results of different methods on the MSR-VTT dataset. }
		\label{tab:msr-vtt}
		
		\begin{tabular}{lp{1cm}p{1cm}p{1cm}} 
			\hline
			Model                                                                                 & BLEU 4     & METEOR         & CIDEr \\
			\hline
			MP-LSTM(V)\cite{venugopalan2014translating}    & 34.8    & 24.8    & -            \\
			Alto\cite{dong2016early}  &39.2     &\textbf{29.4}   &41.0   \\
			MS-RNN(R)\cite{song2017deterministic}    & \textbf{39.8}    & 26.1    & 40.9         \\
			hLSTMt\cite{song2017hierarchical}        & 37.4    & 26.1    & -         \\
			TDDF(VGG+C3D)\cite{zhang122017task}    & 37.3    & {27.8} & \textbf{43.8}         \\
			Attentional Fusion\cite{hori2017attention} &39.7 &25.7 &40.4\\
			Top-down Visual Saliency\cite{ramanishka2017top} & - & 25.9 & - \\
			\hline
			
			\textbf{BI + BD + SA + VP + HLM} (beam)    & \textbf{39.8}    & 26.4         &43.3         \\
			\hline
		\end{tabular}
	\end{center}
\end{table}
\begin{figure*}[t!]
	\centering
	\includegraphics[width=\linewidth]{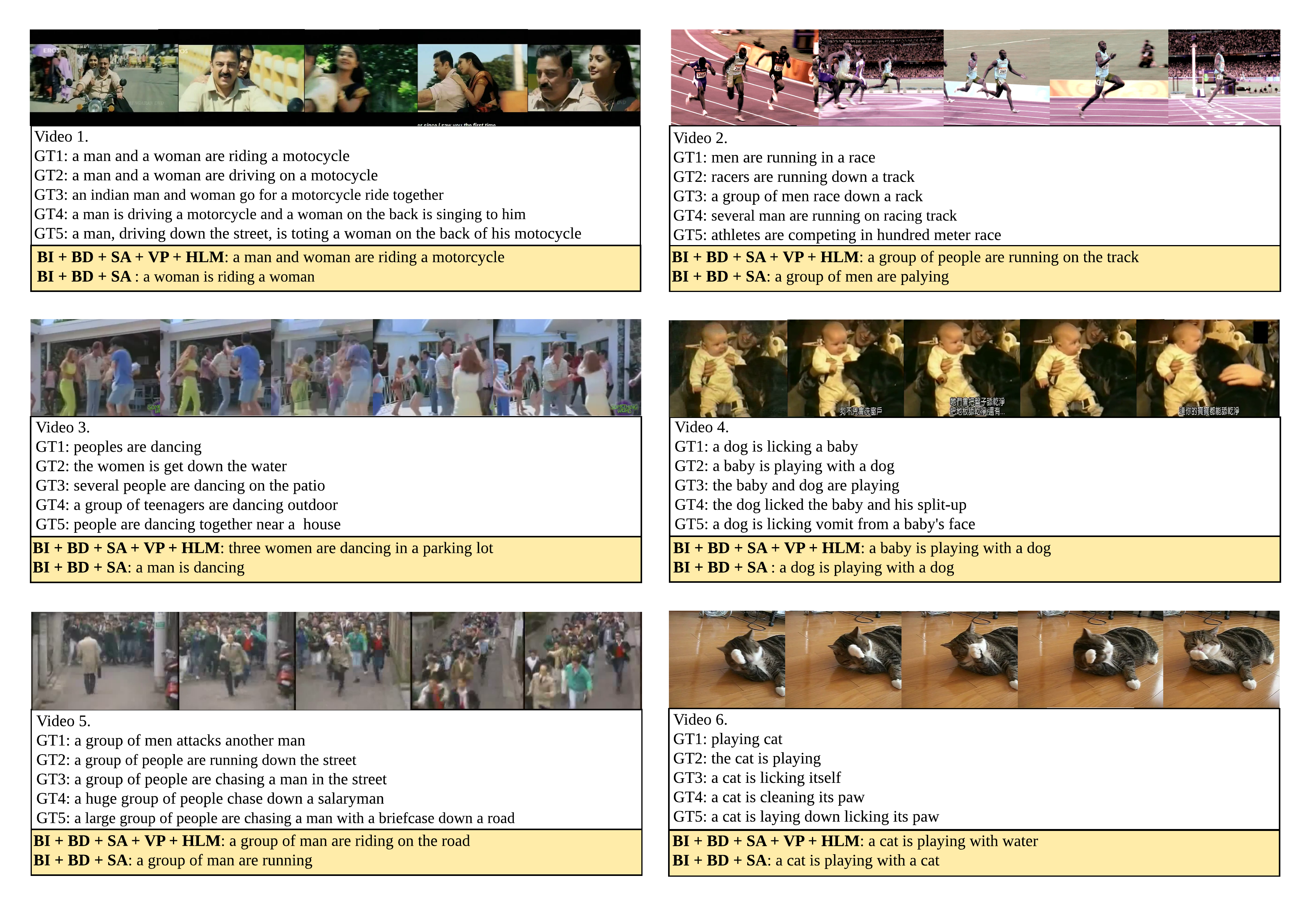}
	\caption{Examples of the generated captions by our final model and a baseline.}
	\label{fig:results}
\end{figure*}

\subsection{Quantitative Results}
Table~\ref{tab:msvd} shows the comparison results of different baselines and the state-of-the-art methods on the MSVD dataset. Comparing among different baselines of our method, we can see that adding each additional component (including BD, SA, VP and HLM) further improves the performance, which demonstrates the effectiveness of each individual component. 
In our model, we share the soft attention module between the video features and the language features, while a common way is to use two different attention modules. To evaluate the effectiveness of the shared attention module, we train our baseline of \textbf{BI + BD + SA + VP} with the two different attention mechanisms (SA$^{\ast}$ using two independent attention modules, and SA using the shared attention module), whose testing results are shown in Table~\ref{tab:msvd}. It can be seen that the proposed shared attention module achieves better performance.
Our final model with beam search, \textbf{BI + BD + SA + VP + HLM}, achieves the best performance in all the baselines across all metrics. Compare with the other approaches trained with the single training dataset, our final model achieves the state-of-the-art performance in BLEU 4 and CIDEr Score.
Table~\ref{tab:msvd} also lists a few other methods that report better results, but all of them make use of multiple datasets for training.

Table~\ref{tab:msr-vtt} shows the video captioning results of different methods on the MSR-VTT dataset. We can see that our final model achieves highly competitive performance, compared with the state-of-the-art methods.

\subsection{Qualitative Results}
Figure~\ref{fig:results} gives some examples of the generated captions.
We can see that our model can avoid some mistakes caused by the conventional language decoding.
Taking the first two videos as example, our model can predict much better results compared with the baseline.
In the fourth video, even though the "dog" is the main object in the video, our model can avoid making the object (e.g, "baby") and the subject (e.g, "dog") to be the same (although our generated caption is also incorrect).
The cases in the fifth and sixth videos are failure cases, which are mainly due to the highly similar objects in the video. Specifically, in the fifth video, our final model confuses ``running" with ``riding" because many people are running in the street and the people behind seem to be "riding" on the front people. In the sixth video, our model predicts the correct subject "cat" but fails to predict the rest words. It is mainly because the reflection of the floor is similar to water.

\begin{figure}[t!]
	\centering
	\includegraphics[width=\linewidth]{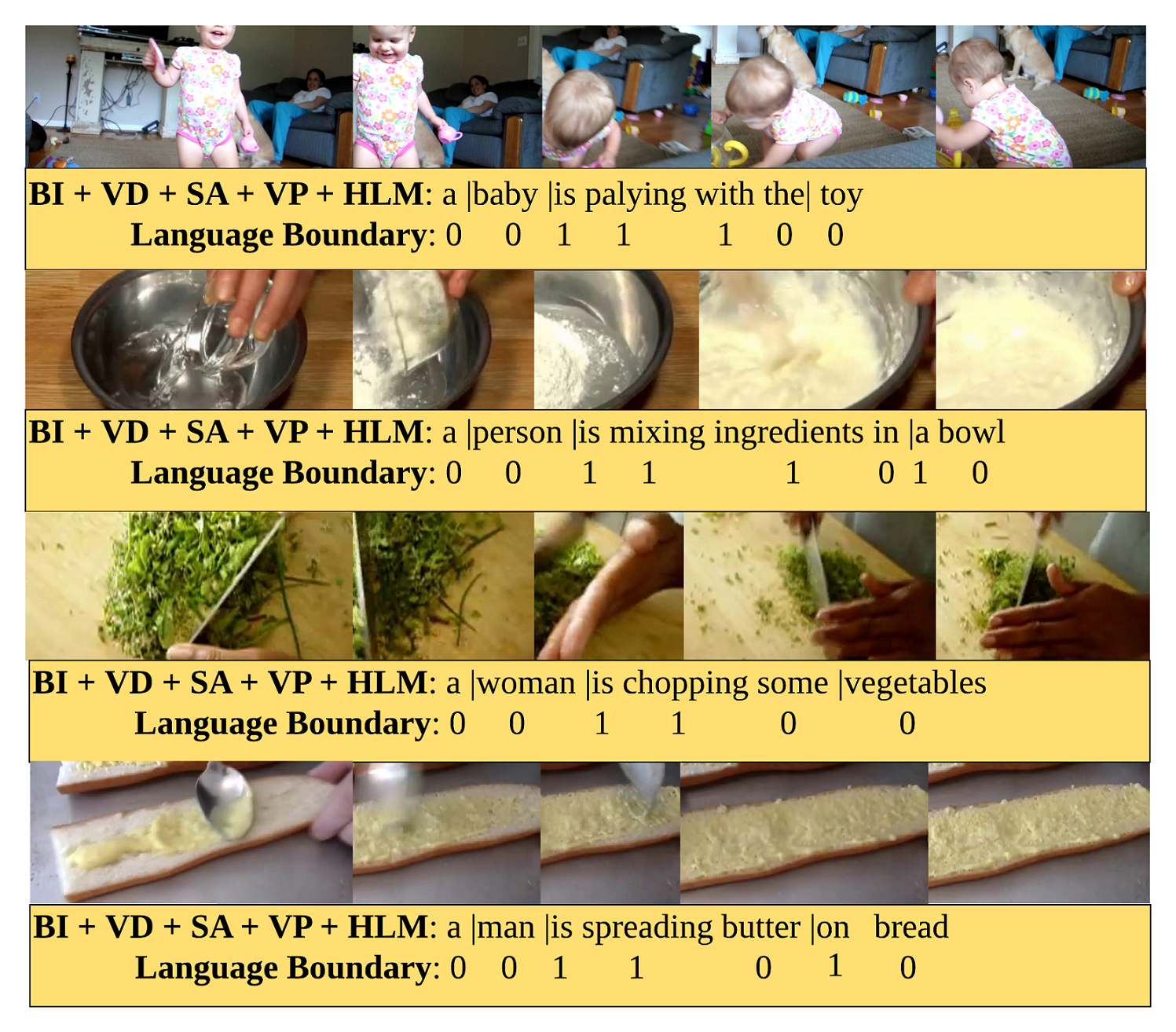}
	\caption{Examples of the language boundary detection results.}
	\label{fig:prove}
\end{figure}

Figure \ref{fig:prove} shows the results of the language boundary detection from the binary gate defined in \eqref{eq:deterministic_step}, where a 0 indicates there is no boundary at that position while 1 indicates there is a boundary. 
We can see that our language boundary detection model can detect the boundary between nouns and verbs. The model considers a sentence as several different word groups such as ``a bowl" and ``on bread". In the first example, that the model considers the word ``the" has a strong dependency with "is playing with" is mainly because the noun after ``the" can change variably, but the word group ``playing with the" has a relatively stable form.


\section{Conclusion}
In this paper, we have proposed a video captioning framework that integrates a boundary-aware hierarchical language decoder, a joint video prediction process, a shared soft attention module, and a boundary-aware video feature encoder. Extensive experiments show that our method can achieve the state-of-the-art performance among the approaches using only a single training dataset. Our proposed boundary-aware hierarchical language model can efficiently predict the language boundaries. In the future, we will extend our method to other tasks such as machine translation and video summary.

\bibliographystyle{ACM-Reference-Format}
\bibliography{sample-bibliography}


\begin{thebibliography}{40}


\ifx \showCODEN    \undefined \def \showCODEN     #1{\unskip}     \fi
\ifx \showDOI      \undefined \def \showDOI       #1{#1}\fi
\ifx \showISBNx    \undefined \def \showISBNx     #1{\unskip}     \fi
\ifx \showISBNxiii \undefined \def \showISBNxiii  #1{\unskip}     \fi
\ifx \showISSN     \undefined \def \showISSN      #1{\unskip}     \fi
\ifx \showLCCN     \undefined \def \showLCCN      #1{\unskip}     \fi
\ifx \shownote     \undefined \def \shownote      #1{#1}          \fi
\ifx \showarticletitle \undefined \def \showarticletitle #1{#1}   \fi
\ifx \showURL      \undefined \def \showURL       {\relax}        \fi
\providecommand\bibfield[2]{#2}
\providecommand\bibinfo[2]{#2}
\providecommand\natexlab[1]{#1}
\providecommand\showeprint[2][]{arXiv:#2}

\bibitem[\protect\citeauthoryear{Baraldi, Grana, and Cucchiara}{Baraldi
  et~al\mbox{.}}{2016}]%
        {baraldi2016hierarchical}
\bibfield{author}{\bibinfo{person}{Lorenzo Baraldi},
  \bibinfo{person}{Costantino Grana}, {and} \bibinfo{person}{Rita Cucchiara}.}
  \bibinfo{year}{2016}\natexlab{}.
\newblock \showarticletitle{Hierarchical Boundary-Aware Neural Encoder for
  Video Captioning}.
\newblock \bibinfo{journal}{\emph{arXiv preprint arXiv:1611.09312}}
  (\bibinfo{year}{2016}).
\newblock


\bibitem[\protect\citeauthoryear{Baraldi, Grana, and Cucchiara}{Baraldi
  et~al\mbox{.}}{2017}]%
        {baraldi2017hierarchical}
\bibfield{author}{\bibinfo{person}{Lorenzo Baraldi},
  \bibinfo{person}{Costantino Grana}, {and} \bibinfo{person}{Rita Cucchiara}.}
  \bibinfo{year}{2017}\natexlab{}.
\newblock \showarticletitle{Hierarchical boundary-aware neural encoder for
  video captioning}. In \bibinfo{booktitle}{\emph{Computer Vision and Pattern
  Recognition (CVPR), 2017 IEEE Conference on}}. IEEE,
  \bibinfo{pages}{3185--3194}.
\newblock


\bibitem[\protect\citeauthoryear{Chen and Dolan}{Chen and Dolan}{2011}]%
        {chen2011collecting}
\bibfield{author}{\bibinfo{person}{David~L Chen} {and}
  \bibinfo{person}{William~B Dolan}.} \bibinfo{year}{2011}\natexlab{}.
\newblock \showarticletitle{Collecting highly parallel data for paraphrase
  evaluation}. In \bibinfo{booktitle}{\emph{Proceedings of the 49th Annual
  Meeting of the Association for Computational Linguistics: Human Language
  Technologies-Volume 1}}. Association for Computational Linguistics,
  \bibinfo{pages}{190--200}.
\newblock


\bibitem[\protect\citeauthoryear{Cho, Van~Merri{\"e}nboer, Gulcehre, Bahdanau,
  Bougares, Schwenk, and Bengio}{Cho et~al\mbox{.}}{2014}]%
        {cho2014learning}
\bibfield{author}{\bibinfo{person}{Kyunghyun Cho}, \bibinfo{person}{Bart
  Van~Merri{\"e}nboer}, \bibinfo{person}{Caglar Gulcehre},
  \bibinfo{person}{Dzmitry Bahdanau}, \bibinfo{person}{Fethi Bougares},
  \bibinfo{person}{Holger Schwenk}, {and} \bibinfo{person}{Yoshua Bengio}.}
  \bibinfo{year}{2014}\natexlab{}.
\newblock \showarticletitle{Learning phrase representations using RNN
  encoder-decoder for statistical machine translation}.
\newblock \bibinfo{journal}{\emph{EMNLP}} (\bibinfo{year}{2014}).
\newblock


\bibitem[\protect\citeauthoryear{Chung, Gulcehre, Cho, and Bengio}{Chung
  et~al\mbox{.}}{2014}]%
        {chung2014empirical}
\bibfield{author}{\bibinfo{person}{Junyoung Chung}, \bibinfo{person}{Caglar
  Gulcehre}, \bibinfo{person}{KyungHyun Cho}, {and} \bibinfo{person}{Yoshua
  Bengio}.} \bibinfo{year}{2014}\natexlab{}.
\newblock \showarticletitle{Empirical evaluation of gated recurrent neural
  networks on sequence modeling}.
\newblock \bibinfo{journal}{\emph{arXiv preprint arXiv:1412.3555}}
  (\bibinfo{year}{2014}).
\newblock


\bibitem[\protect\citeauthoryear{Deng, Dong, Socher, Li, Li, and Fei-Fei}{Deng
  et~al\mbox{.}}{2009}]%
        {deng2009imagenet}
\bibfield{author}{\bibinfo{person}{Jia Deng}, \bibinfo{person}{Wei Dong},
  \bibinfo{person}{Richard Socher}, \bibinfo{person}{Li-Jia Li},
  \bibinfo{person}{Kai Li}, {and} \bibinfo{person}{Li Fei-Fei}.}
  \bibinfo{year}{2009}\natexlab{}.
\newblock \showarticletitle{Imagenet: A large-scale hierarchical image
  database}. In \bibinfo{booktitle}{\emph{Computer Vision and Pattern
  Recognition, 2009. CVPR 2009. IEEE Conference on}}. IEEE,
  \bibinfo{pages}{248--255}.
\newblock


\bibitem[\protect\citeauthoryear{Denkowski and Lavie}{Denkowski and
  Lavie}{2014}]%
        {denkowski2014meteor}
\bibfield{author}{\bibinfo{person}{Michael Denkowski} {and}
  \bibinfo{person}{Alon Lavie}.} \bibinfo{year}{2014}\natexlab{}.
\newblock \showarticletitle{Meteor universal: Language specific translation
  evaluation for any target language}. In \bibinfo{booktitle}{\emph{ACL}}.
\newblock


\bibitem[\protect\citeauthoryear{Donahue, Anne~Hendricks, Guadarrama, Rohrbach,
  Venugopalan, Saenko, and Darrell}{Donahue et~al\mbox{.}}{2015}]%
        {donahue2015long}
\bibfield{author}{\bibinfo{person}{Jeffrey Donahue}, \bibinfo{person}{Lisa
  Anne~Hendricks}, \bibinfo{person}{Sergio Guadarrama}, \bibinfo{person}{Marcus
  Rohrbach}, \bibinfo{person}{Subhashini Venugopalan}, \bibinfo{person}{Kate
  Saenko}, {and} \bibinfo{person}{Trevor Darrell}.}
  \bibinfo{year}{2015}\natexlab{}.
\newblock \showarticletitle{Long-term recurrent convolutional networks for
  visual recognition and description}. In \bibinfo{booktitle}{\emph{CVPR}}.
\newblock


\bibitem[\protect\citeauthoryear{Dong, Li, Lan, Huo, and Snoek}{Dong
  et~al\mbox{.}}{2016}]%
        {dong2016early}
\bibfield{author}{\bibinfo{person}{Jianfeng Dong}, \bibinfo{person}{Xirong Li},
  \bibinfo{person}{Weiyu Lan}, \bibinfo{person}{Yujia Huo}, {and}
  \bibinfo{person}{Cees~GM Snoek}.} \bibinfo{year}{2016}\natexlab{}.
\newblock \showarticletitle{Early embedding and late reranking for video
  captioning}. In \bibinfo{booktitle}{\emph{Proceedings of the 2016 ACM on
  Multimedia Conference}}. ACM, \bibinfo{pages}{1082--1086}.
\newblock


\bibitem[\protect\citeauthoryear{Gan, Gan, He, Pu, Tran, Gao, Carin, and
  Deng}{Gan et~al\mbox{.}}{2017}]%
        {gan2017semantic}
\bibfield{author}{\bibinfo{person}{Zhe Gan}, \bibinfo{person}{Chuang Gan},
  \bibinfo{person}{Xiaodong He}, \bibinfo{person}{Yunchen Pu},
  \bibinfo{person}{Kenneth Tran}, \bibinfo{person}{Jianfeng Gao},
  \bibinfo{person}{Lawrence Carin}, {and} \bibinfo{person}{Li Deng}.}
  \bibinfo{year}{2017}\natexlab{}.
\newblock \showarticletitle{Semantic compositional networks for visual
  captioning}. In \bibinfo{booktitle}{\emph{Proceedings of the IEEE Conference
  on Computer Vision and Pattern Recognition}}, Vol.~\bibinfo{volume}{2}.
\newblock


\bibitem[\protect\citeauthoryear{Guadarrama, Krishnamoorthy, Malkarnenkar,
  Venugopalan, Mooney, Darrell, and Saenko}{Guadarrama et~al\mbox{.}}{2013}]%
        {guadarrama2013youtube2text}
\bibfield{author}{\bibinfo{person}{Sergio Guadarrama}, \bibinfo{person}{Niveda
  Krishnamoorthy}, \bibinfo{person}{Girish Malkarnenkar},
  \bibinfo{person}{Subhashini Venugopalan}, \bibinfo{person}{Raymond Mooney},
  \bibinfo{person}{Trevor Darrell}, {and} \bibinfo{person}{Kate Saenko}.}
  \bibinfo{year}{2013}\natexlab{}.
\newblock \showarticletitle{Youtube2text: Recognizing and describing arbitrary
  activities using semantic hierarchies and zero-shot recognition}. In
  \bibinfo{booktitle}{\emph{Computer Vision (ICCV), 2013 IEEE International
  Conference on}}. IEEE, \bibinfo{pages}{2712--2719}.
\newblock


\bibitem[\protect\citeauthoryear{He, Zhang, Ren, and Sun}{He
  et~al\mbox{.}}{2016}]%
        {he2016deep}
\bibfield{author}{\bibinfo{person}{Kaiming He}, \bibinfo{person}{Xiangyu
  Zhang}, \bibinfo{person}{Shaoqing Ren}, {and} \bibinfo{person}{Jian Sun}.}
  \bibinfo{year}{2016}\natexlab{}.
\newblock \showarticletitle{Deep residual learning for image recognition}. In
  \bibinfo{booktitle}{\emph{Proceedings of the IEEE conference on computer
  vision and pattern recognition}}. \bibinfo{pages}{770--778}.
\newblock


\bibitem[\protect\citeauthoryear{Hochreiter and Schmidhuber}{Hochreiter and
  Schmidhuber}{1997}]%
        {hochreiter1997long}
\bibfield{author}{\bibinfo{person}{Sepp Hochreiter} {and}
  \bibinfo{person}{J{\"u}rgen Schmidhuber}.} \bibinfo{year}{1997}\natexlab{}.
\newblock \showarticletitle{Long short-term memory}.
\newblock \bibinfo{journal}{\emph{Neural computation}} (\bibinfo{year}{1997}).
\newblock


\bibitem[\protect\citeauthoryear{Hori, Hori, Lee, Sumi, Hershey, and
  Marks}{Hori et~al\mbox{.}}{2017}]%
        {hori2017attention}
\bibfield{author}{\bibinfo{person}{Chiori Hori}, \bibinfo{person}{Takaaki
  Hori}, \bibinfo{person}{Teng-Yok Lee}, \bibinfo{person}{Kazuhiro Sumi},
  \bibinfo{person}{John~R Hershey}, {and} \bibinfo{person}{Tim~K Marks}.}
  \bibinfo{year}{2017}\natexlab{}.
\newblock \showarticletitle{Attention-based multimodal fusion for video
  description}.
\newblock \bibinfo{journal}{\emph{arXiv preprint}} (\bibinfo{year}{2017}).
\newblock


\bibitem[\protect\citeauthoryear{Kang, Zhang, and Liu}{Kang
  et~al\mbox{.}}{2016}]%
        {kang2016gated}
\bibfield{author}{\bibinfo{person}{Jian Kang}, \bibinfo{person}{Wei-Qiang
  Zhang}, {and} \bibinfo{person}{Jia Liu}.} \bibinfo{year}{2016}\natexlab{}.
\newblock \showarticletitle{Gated recurrent units based hybrid acoustic models
  for robust speech recognition}. In \bibinfo{booktitle}{\emph{Chinese Spoken
  Language Processing (ISCSLP), 2016 10th International Symposium on}}. IEEE,
  \bibinfo{pages}{1--5}.
\newblock


\bibitem[\protect\citeauthoryear{Karpathy and Fei-Fei}{Karpathy and
  Fei-Fei}{2015}]%
        {karpathy2015deep}
\bibfield{author}{\bibinfo{person}{Andrej Karpathy} {and} \bibinfo{person}{Li
  Fei-Fei}.} \bibinfo{year}{2015}\natexlab{}.
\newblock \showarticletitle{Deep visual-semantic alignments for generating
  image descriptions}. In \bibinfo{booktitle}{\emph{CVPR}}.
\newblock


\bibitem[\protect\citeauthoryear{Kingma and Ba}{Kingma and Ba}{2015}]%
        {kingma2014adam}
\bibfield{author}{\bibinfo{person}{Diederik Kingma} {and}
  \bibinfo{person}{Jimmy Ba}.} \bibinfo{year}{2015}\natexlab{}.
\newblock \showarticletitle{Adam: A method for stochastic optimization}.
\newblock \bibinfo{journal}{\emph{ICLR}} (\bibinfo{year}{2015}).
\newblock


\bibitem[\protect\citeauthoryear{Krishnamoorthy, Malkarnenkar, Mooney, Saenko,
  and Guadarrama}{Krishnamoorthy et~al\mbox{.}}{2013}]%
        {krishnamoorthy2013generating}
\bibfield{author}{\bibinfo{person}{Niveda Krishnamoorthy},
  \bibinfo{person}{Girish Malkarnenkar}, \bibinfo{person}{Raymond~J Mooney},
  \bibinfo{person}{Kate Saenko}, {and} \bibinfo{person}{Sergio Guadarrama}.}
  \bibinfo{year}{2013}\natexlab{}.
\newblock \showarticletitle{Generating Natural-Language Video Descriptions
  Using Text-Mined Knowledge.}. In \bibinfo{booktitle}{\emph{AAAI}},
  Vol.~\bibinfo{volume}{1}. \bibinfo{pages}{2}.
\newblock


\bibitem[\protect\citeauthoryear{Luong, Le, Sutskever, Vinyals, and
  Kaiser}{Luong et~al\mbox{.}}{2015}]%
        {luong2015multi}
\bibfield{author}{\bibinfo{person}{Minh-Thang Luong}, \bibinfo{person}{Quoc~V
  Le}, \bibinfo{person}{Ilya Sutskever}, \bibinfo{person}{Oriol Vinyals}, {and}
  \bibinfo{person}{Lukasz Kaiser}.} \bibinfo{year}{2015}\natexlab{}.
\newblock \showarticletitle{Multi-task sequence to sequence learning}.
\newblock \bibinfo{journal}{\emph{arXiv preprint arXiv:1511.06114}}
  (\bibinfo{year}{2015}).
\newblock


\bibitem[\protect\citeauthoryear{Pan, Xu, Yang, Wu, and Zhuang}{Pan
  et~al\mbox{.}}{2016b}]%
        {pan2016hierarchical}
\bibfield{author}{\bibinfo{person}{Pingbo Pan}, \bibinfo{person}{Zhongwen Xu},
  \bibinfo{person}{Yi Yang}, \bibinfo{person}{Fei Wu}, {and}
  \bibinfo{person}{Yueting Zhuang}.} \bibinfo{year}{2016}\natexlab{b}.
\newblock \showarticletitle{Hierarchical recurrent neural encoder for video
  representation with application to captioning}. In
  \bibinfo{booktitle}{\emph{Proceedings of the IEEE Conference on Computer
  Vision and Pattern Recognition}}. \bibinfo{pages}{1029--1038}.
\newblock


\bibitem[\protect\citeauthoryear{Pan, Mei, Yao, Li, and Rui}{Pan
  et~al\mbox{.}}{2016a}]%
        {pan2016jointly}
\bibfield{author}{\bibinfo{person}{Yingwei Pan}, \bibinfo{person}{Tao Mei},
  \bibinfo{person}{Ting Yao}, \bibinfo{person}{Houqiang Li}, {and}
  \bibinfo{person}{Yong Rui}.} \bibinfo{year}{2016}\natexlab{a}.
\newblock \showarticletitle{Jointly modeling embedding and translation to
  bridge video and language}. In \bibinfo{booktitle}{\emph{Proceedings of the
  IEEE conference on computer vision and pattern recognition}}.
  \bibinfo{pages}{4594--4602}.
\newblock


\bibitem[\protect\citeauthoryear{Pan, Yao, Li, and Mei}{Pan
  et~al\mbox{.}}{2017}]%
        {pan2017video}
\bibfield{author}{\bibinfo{person}{Yingwei Pan}, \bibinfo{person}{Ting Yao},
  \bibinfo{person}{Houqiang Li}, {and} \bibinfo{person}{Tao Mei}.}
  \bibinfo{year}{2017}\natexlab{}.
\newblock \showarticletitle{Video captioning with transferred semantic
  attributes}. In \bibinfo{booktitle}{\emph{CVPR}}.
\newblock


\bibitem[\protect\citeauthoryear{Papineni, Roukos, Ward, and Zhu}{Papineni
  et~al\mbox{.}}{2002}]%
        {papineni2002bleu}
\bibfield{author}{\bibinfo{person}{Kishore Papineni}, \bibinfo{person}{Salim
  Roukos}, \bibinfo{person}{Todd Ward}, {and} \bibinfo{person}{Wei-Jing Zhu}.}
  \bibinfo{year}{2002}\natexlab{}.
\newblock \showarticletitle{BLEU: a method for automatic evaluation of machine
  translation}. In \bibinfo{booktitle}{\emph{ACL}}.
\newblock


\bibitem[\protect\citeauthoryear{Pasunuru and Bansal}{Pasunuru and
  Bansal}{2017}]%
        {pasunuru2017multi}
\bibfield{author}{\bibinfo{person}{Ramakanth Pasunuru} {and}
  \bibinfo{person}{Mohit Bansal}.} \bibinfo{year}{2017}\natexlab{}.
\newblock \showarticletitle{Multi-Task Video Captioning with Video and
  Entailment Generation}.
\newblock \bibinfo{journal}{\emph{arXiv preprint arXiv:1704.07489}}
  (\bibinfo{year}{2017}).
\newblock


\bibitem[\protect\citeauthoryear{Ramanishka, Das, Zhang, and Saenko}{Ramanishka
  et~al\mbox{.}}{2017}]%
        {ramanishka2017top}
\bibfield{author}{\bibinfo{person}{Vasili Ramanishka}, \bibinfo{person}{Abir
  Das}, \bibinfo{person}{Jianming Zhang}, {and} \bibinfo{person}{Kate Saenko}.}
  \bibinfo{year}{2017}\natexlab{}.
\newblock \showarticletitle{Top-down visual saliency guided by captions}. In
  \bibinfo{booktitle}{\emph{Proceedings of the IEEE Conference on Computer
  Vision and Pattern Recognition (CVPR)}}, Vol.~\bibinfo{volume}{1}.
  \bibinfo{pages}{7}.
\newblock


\bibitem[\protect\citeauthoryear{Rohrbach, Rohrbach, and Schiele}{Rohrbach
  et~al\mbox{.}}{2015}]%
        {rohrbach2015long}
\bibfield{author}{\bibinfo{person}{Anna Rohrbach}, \bibinfo{person}{Marcus
  Rohrbach}, {and} \bibinfo{person}{Bernt Schiele}.}
  \bibinfo{year}{2015}\natexlab{}.
\newblock \showarticletitle{The long-short story of movie description}. In
  \bibinfo{booktitle}{\emph{German Conference on Pattern Recognition}}.
  Springer, \bibinfo{pages}{209--221}.
\newblock


\bibitem[\protect\citeauthoryear{Song, Guo, Gao, Li, Hanjalic, and Shen}{Song
  et~al\mbox{.}}{2017a}]%
        {song2017deterministic}
\bibfield{author}{\bibinfo{person}{Jingkuan Song}, \bibinfo{person}{Yuyu Guo},
  \bibinfo{person}{Lianli Gao}, \bibinfo{person}{Xuelong Li},
  \bibinfo{person}{Alan Hanjalic}, {and} \bibinfo{person}{Heng~Tao Shen}.}
  \bibinfo{year}{2017}\natexlab{a}.
\newblock \showarticletitle{From Deterministic to Generative: Multi-Modal
  Stochastic RNNs for Video Captioning}.
\newblock \bibinfo{journal}{\emph{arXiv preprint arXiv:1708.02478}}
  (\bibinfo{year}{2017}).
\newblock


\bibitem[\protect\citeauthoryear{Song, Guo, Gao, Liu, Zhang, and Shen}{Song
  et~al\mbox{.}}{2017b}]%
        {song2017hierarchical}
\bibfield{author}{\bibinfo{person}{Jingkuan Song}, \bibinfo{person}{Zhao Guo},
  \bibinfo{person}{Lianli Gao}, \bibinfo{person}{Wu Liu},
  \bibinfo{person}{Dongxiang Zhang}, {and} \bibinfo{person}{Heng~Tao Shen}.}
  \bibinfo{year}{2017}\natexlab{b}.
\newblock \showarticletitle{Hierarchical lstm with adjusted temporal attention
  for video captioning}.
\newblock \bibinfo{journal}{\emph{arXiv preprint arXiv:1706.01231}}
  (\bibinfo{year}{2017}).
\newblock


\bibitem[\protect\citeauthoryear{Soomro, Zamir, and Shah}{Soomro
  et~al\mbox{.}}{2012}]%
        {soomro2012ucf101}
\bibfield{author}{\bibinfo{person}{Khurram Soomro},
  \bibinfo{person}{Amir~Roshan Zamir}, {and} \bibinfo{person}{Mubarak Shah}.}
  \bibinfo{year}{2012}\natexlab{}.
\newblock \showarticletitle{UCF101: A dataset of 101 human actions classes from
  videos in the wild}.
\newblock \bibinfo{journal}{\emph{arXiv preprint arXiv:1212.0402}}
  (\bibinfo{year}{2012}).
\newblock


\bibitem[\protect\citeauthoryear{Sutskever, Vinyals, and Le}{Sutskever
  et~al\mbox{.}}{2014}]%
        {sutskever2014sequence}
\bibfield{author}{\bibinfo{person}{Ilya Sutskever}, \bibinfo{person}{Oriol
  Vinyals}, {and} \bibinfo{person}{Quoc~V Le}.}
  \bibinfo{year}{2014}\natexlab{}.
\newblock \showarticletitle{Sequence to sequence learning with neural
  networks}. In \bibinfo{booktitle}{\emph{NIPS}}.
\newblock


\bibitem[\protect\citeauthoryear{Thomason, Venugopalan, Guadarrama, Saenko, and
  Mooney}{Thomason et~al\mbox{.}}{2014}]%
        {thomason2014integrating}
\bibfield{author}{\bibinfo{person}{Jesse Thomason}, \bibinfo{person}{Subhashini
  Venugopalan}, \bibinfo{person}{Sergio Guadarrama}, \bibinfo{person}{Kate
  Saenko}, {and} \bibinfo{person}{Raymond~J Mooney}.}
  \bibinfo{year}{2014}\natexlab{}.
\newblock \showarticletitle{Integrating Language and Vision to Generate Natural
  Language Descriptions of Videos in the Wild.}. In
  \bibinfo{booktitle}{\emph{Coling}}, Vol.~\bibinfo{volume}{2}.
  \bibinfo{pages}{9}.
\newblock


\bibitem[\protect\citeauthoryear{Tran, Bourdev, Fergus, Torresani, and
  Paluri}{Tran et~al\mbox{.}}{2015}]%
        {tran2015learning}
\bibfield{author}{\bibinfo{person}{Du Tran}, \bibinfo{person}{Lubomir Bourdev},
  \bibinfo{person}{Rob Fergus}, \bibinfo{person}{Lorenzo Torresani}, {and}
  \bibinfo{person}{Manohar Paluri}.} \bibinfo{year}{2015}\natexlab{}.
\newblock \showarticletitle{Learning spatiotemporal features with 3d
  convolutional networks}. In \bibinfo{booktitle}{\emph{Computer Vision (ICCV),
  2015 IEEE International Conference on}}. IEEE, \bibinfo{pages}{4489--4497}.
\newblock


\bibitem[\protect\citeauthoryear{Vedantam, Lawrence~Zitnick, and
  Parikh}{Vedantam et~al\mbox{.}}{2015}]%
        {vedantam2015cider}
\bibfield{author}{\bibinfo{person}{Ramakrishna Vedantam}, \bibinfo{person}{C
  Lawrence~Zitnick}, {and} \bibinfo{person}{Devi Parikh}.}
  \bibinfo{year}{2015}\natexlab{}.
\newblock \showarticletitle{Cider: Consensus-based image description
  evaluation}. In \bibinfo{booktitle}{\emph{CVPR}}.
\newblock


\bibitem[\protect\citeauthoryear{Venugopalan, Rohrbach, Donahue, Mooney,
  Darrell, and Saenko}{Venugopalan et~al\mbox{.}}{2015}]%
        {venugopalan2015sequence}
\bibfield{author}{\bibinfo{person}{Subhashini Venugopalan},
  \bibinfo{person}{Marcus Rohrbach}, \bibinfo{person}{Jeffrey Donahue},
  \bibinfo{person}{Raymond Mooney}, \bibinfo{person}{Trevor Darrell}, {and}
  \bibinfo{person}{Kate Saenko}.} \bibinfo{year}{2015}\natexlab{}.
\newblock \showarticletitle{Sequence to sequence-video to text}. In
  \bibinfo{booktitle}{\emph{Proceedings of the IEEE international conference on
  computer vision}}. \bibinfo{pages}{4534--4542}.
\newblock


\bibitem[\protect\citeauthoryear{Venugopalan, Xu, Donahue, Rohrbach, Mooney,
  and Saenko}{Venugopalan et~al\mbox{.}}{2014}]%
        {venugopalan2014translating}
\bibfield{author}{\bibinfo{person}{Subhashini Venugopalan},
  \bibinfo{person}{Huijuan Xu}, \bibinfo{person}{Jeff Donahue},
  \bibinfo{person}{Marcus Rohrbach}, \bibinfo{person}{Raymond Mooney}, {and}
  \bibinfo{person}{Kate Saenko}.} \bibinfo{year}{2014}\natexlab{}.
\newblock \showarticletitle{Translating videos to natural language using deep
  recurrent neural networks}.
\newblock \bibinfo{journal}{\emph{arXiv preprint arXiv:1412.4729}}
  (\bibinfo{year}{2014}).
\newblock


\bibitem[\protect\citeauthoryear{Vinyals, Toshev, Bengio, and Erhan}{Vinyals
  et~al\mbox{.}}{2015}]%
        {vinyals2015show}
\bibfield{author}{\bibinfo{person}{Oriol Vinyals}, \bibinfo{person}{Alexander
  Toshev}, \bibinfo{person}{Samy Bengio}, {and} \bibinfo{person}{Dumitru
  Erhan}.} \bibinfo{year}{2015}\natexlab{}.
\newblock \showarticletitle{Show and tell: A neural image caption generator}.
  In \bibinfo{booktitle}{\emph{CVPR}}.
\newblock


\bibitem[\protect\citeauthoryear{Xu, Venugopalan, Ramanishka, Rohrbach, and
  Saenko}{Xu et~al\mbox{.}}{2015}]%
        {xu2015multi}
\bibfield{author}{\bibinfo{person}{Huijuan Xu}, \bibinfo{person}{Subhashini
  Venugopalan}, \bibinfo{person}{Vasili Ramanishka}, \bibinfo{person}{Marcus
  Rohrbach}, {and} \bibinfo{person}{Kate Saenko}.}
  \bibinfo{year}{2015}\natexlab{}.
\newblock \showarticletitle{A multi-scale multiple instance video description
  network}.
\newblock \bibinfo{journal}{\emph{arXiv preprint arXiv:1505.05914}}
  (\bibinfo{year}{2015}).
\newblock


\bibitem[\protect\citeauthoryear{Xu, Mei, Yao, and Rui}{Xu
  et~al\mbox{.}}{2016}]%
        {xu2016msr}
\bibfield{author}{\bibinfo{person}{Jun Xu}, \bibinfo{person}{Tao Mei},
  \bibinfo{person}{Ting Yao}, {and} \bibinfo{person}{Yong Rui}.}
  \bibinfo{year}{2016}\natexlab{}.
\newblock \showarticletitle{Msr-vtt: A large video description dataset for
  bridging video and language}. In \bibinfo{booktitle}{\emph{Computer Vision
  and Pattern Recognition (CVPR), 2016 IEEE Conference on}}. IEEE,
  \bibinfo{pages}{5288--5296}.
\newblock


\bibitem[\protect\citeauthoryear{Yao, Torabi, Cho, Ballas, Pal, Larochelle, and
  Courville}{Yao et~al\mbox{.}}{2015}]%
        {yao2015describing}
\bibfield{author}{\bibinfo{person}{Li Yao}, \bibinfo{person}{Atousa Torabi},
  \bibinfo{person}{Kyunghyun Cho}, \bibinfo{person}{Nicolas Ballas},
  \bibinfo{person}{Christopher Pal}, \bibinfo{person}{Hugo Larochelle}, {and}
  \bibinfo{person}{Aaron Courville}.} \bibinfo{year}{2015}\natexlab{}.
\newblock \showarticletitle{Describing videos by exploiting temporal
  structure}. In \bibinfo{booktitle}{\emph{Proceedings of the IEEE
  international conference on computer vision}}. \bibinfo{pages}{4507--4515}.
\newblock


\bibitem[\protect\citeauthoryear{Zhang12, Gao, Zhang12, Zhang, Li, and
  Tian}{Zhang12 et~al\mbox{.}}{2017}]%
        {zhang122017task}
\bibfield{author}{\bibinfo{person}{Xishan Zhang12}, \bibinfo{person}{Ke Gao},
  \bibinfo{person}{Yongdong Zhang12}, \bibinfo{person}{Dongming Zhang},
  \bibinfo{person}{Jintao Li}, {and} \bibinfo{person}{Qi Tian}.}
  \bibinfo{year}{2017}\natexlab{}.
\newblock \showarticletitle{Task-Driven Dynamic Fusion: Reducing Ambiguity in
  Video Description}.
\newblock  (\bibinfo{year}{2017}).
\newblock


\end{thebibliography}

\end{document}